\documentclass{article}

\usepackage{microtype}
\usepackage{graphicx}
\usepackage{subfigure}
\usepackage{booktabs} 

\usepackage{hyperref}

\usepackage[accepted]{icml2025}

\usepackage{amsmath}
\usepackage{amssymb}
\usepackage{mathtools}
\usepackage{amsthm}

\usepackage[capitalize,noabbrev]{cleveref}

\usepackage{icml2025}

\theoremstyle{plain}

\theoremstyle{definition}

\theoremstyle{remark}

\usepackage[textsize=tiny]{todonotes}

\icmltitlerunning{Stop Acting Like Language Model Agents Are Normal Agents}

\begin{document}
\twocolumn
\icmltitle{Position: Stop Acting Like Language Model Agents Are Normal Agents}




\begin{icmlauthorlist}
\icmlauthor{Elija Perrier}{comp,equal,yyy}
\icmlauthor{Michael Timothy Bennett}{equal}
\end{icmlauthorlist}

\icmlaffiliation{yyy}{Centre for Quantum Software and Information, UTS, Sydney}
\icmlaffiliation{equal}{Australian National University, Canberra, Australia}
\icmlaffiliation{comp}{Stanford Center for Responsible Quantum Technology, Stanford University, United States}

\icmlcorrespondingauthor{Elija Perrier}{elija dot perrier at gmail dot com}

\icmlkeywords{Agents, Identity, Ontology}

\vskip 0.3in



\printAffiliationsAndNotice{}  

\begin{abstract}
Language Model Agents (LMAs) are increasingly treated as capable of autonomously navigating interactions with humans and tools. Their design and deployment tends to presume they are normal agents capable of sustaining coherent goals, adapting across contexts and acting with a measure of intentionality. These assumptions are critical to prospective use cases in industrial, social and governmental settings. But LMAs are not normal agents. They inherit the structural problems of the large language models (LLMs) around which they are built: hallucinations, jailbreaking, misalignment and unpredictability. In this Position paper we argue LMAs should not be treated as normal agents, because doing so leads to problems that undermine their utility and trustworthiness. 
We enumerate pathologies of agency intrinsic to LMAs. Despite scaffolding such as external memory and tools, they remain ontologically stateless, stochastic, semantically sensitive, and linguistically intermediated. These pathologies destabilise the ontological properties of LMAs including identifiability, continuity, persistence and and consistency, problematising their claim to agency. In response, we argue LMA ontological properties should be measured before, during and after deployment so that the negative effects of pathologies can be mitigated.
\end{abstract}

\section{Introduction}
Language Model Agents (LMAs) \cite{xi2023rise} are agentic systems based upon large language models (LLMs). They appear to be able to reason \cite{wei_chain--thought_2023} and plan \cite{innermonologue,momennejad_evaluating_2023} in natural language, which allows them to interact autonomously \cite{kinniment_evaluating_2023} with a wide range of human systems. This overcomes some persistent limitations of classical agents \cite{bennettmaruyama2022b}. LMAs are already being used in finance \cite{han2024enhancing,zhou2024finrobot,yu2023finmem,zhang2024stockagent}, politics \cite{Yin2023-mf}, software engineering \cite{chowdhury_introducing_2024,jimenez_swe-bench_2024}, healthcare \cite{schmidgall2024agentclinic,Mehandru_Miao_Almaraz_Sushil_Butte_Alaa_2024,Tu_Palepu_Schaekermann_Saab_Freyberg_Tanno_Wang_Li_Amin_Tomasev_etal._2024}, customer service, legal services and insurance for claims management. They are being talked about as the bedrock of new economies \cite{weforum2024ai,reuters2024autonomous} in which autonomous or semi-autonomous agents transact \cite{wsj2024aiagents}, negotiate, organise and act at scale and speed. As a result, frontier AI laboratories \cite{altman_ai_killer_function} and technology companies \cite{venturebeat2024microsoft_ai_agents,fortune2025nvidia} are becoming increasingly focused on the use of LMAs \cite{altman_level3_agents,pwc2024agentic,forbes2024agentic_ai,forbes2024navigating_shift}. 

LMAs are expected by some become the default interface or `control layer' between humans and the cyberphysical systems with which we interact \cite{a16z_control_layer,ganapathy2021,bennett2025a}. Agency is a necessary step in the direction of artificial general intelligence \cite{goertzel2021,thorisson2012,wang2006rigid}. If an LMA acts like an agent, humans will tend to treat it as if it is \cite{kotrschal2015,bennett2023d}. Yet an LMA might only mimic human-like behaviour \cite{floridi2020,bennettmaruyama2022a}. It may not be agentic in the normal sense \cite{maes_agents_1994,maes_artificial_1995,lieberman_autonomous_1997,jennings_roadmap_1998,johnson_software_2011,sutton_reinforcement_2018,russell_artificial_2021,chan_harms_2023,wu_autogen_2023,openai_openai_2018,gabriel_ethics_2024,kolt_governing_2024,lazar_frontier_2024}. We will argue the extent to which LMAs are truly agentic is a significant factor in how and where they are useful. We identify pathologies of agency intrinsic to LMAs, and argue these pathologies must be acknowledged and measured if they are to be mitigated. We conclude that LMAs are not agents in the \textit{normal} sense, and by acting as if they are we limit their usefulness. Their uncanny nature need not be an impediment.


\subsection{Differentiation from Related Work}
Considerable attention has been paid to how LLMs hallucinate \cite{Rawte_Chakraborty_Pathak_Sarkar_Tonmoy_Chadha_Sheth_Das_2023}, provide incorrect information \cite{Mündler_He_Jenko_2023} or otherwise act inconsistently. Emerging research is starting to address reliability of LMA capabilities \cite{kapoor_ai_2024,DBLP:journals/arxiv/abs-2308-03688,liu_agentbench_2023,mialon_gaia_2023,lu_toolsandbox_2024,zhang_cybench_2024,jimenez_swe-bench_2024,wu_introducing_2024,chowdhury_introducing_2024,gur_real-world_2024,multion_multion_2024,gu2024agent}. But comparatively little research exists on how the nature of LLMs affects the classical \textit{agentic} properties of LMAs upon which their widespread use relies. Most proposals for LMAs assume they identifiable, can be distinguished from their environments, persist continuously over time and act consistently. These assumptions are problematic. LMAs are built upon LLMs which are stateless, stochastic, semantically sensitive and linguistically intermediated. These properties or `pathologies' make identifying continuous, persistent and consistent LMAs and their boundaries difficult. This problematises claims that LMAs reliably satisfy agency criteria. In effect, LMAs suffer from an identity crisis. This has consequences for their reliability, utility and trustworthiness.

\subsection{Position}
Here we argue that LMAs face an identity crisis. This stems from the LLMs around which they are constructed. LMAs are not \textit{normal} agents. They have intrinsic pathologies, which we enumerate here. 
Yet many LMA proposals treat LMAs as something like normal agents. This is both incorrect and normatively undesirable. It engenders a false sense of utility and trustworthiness in LMAs. We do not argue against the use of LMAs. Quite the contrary. Our position is that the identity crisis underpinning LMAs requires attention and, perhaps most importantly, means of scientifically evaluating the extent to which LMAs are ontologically robust and actually satisfy agentic criteria. The pathologies of LMAs should be acknowledged, measured and mitigated. We discuss these issues and sketch out our proposal for agentic evaluation below.



\section{Agentic Identity}
\subsection{What is a Language Model Agent?}
To govern an agent of any sort we need to specify what it is, and is not. We need identity criteria. Entities which satisfy identity criteria (in whole or part) are classified as agents. Theories of agency describe such criteria, providing an ontology of agents: what types of agent there are, the classification of their properties and how they may vary. While there is no universal, agreed upon definition of what constitutes an agent e.g. \cite{franklin1997agent, Schlosser_2019}, there are common features. At minimum, an agent must be distinguished from its environment. It must be capable of action, pursuing goals and interacting with the environment responsively in accordance with plans, practical reasoning or intentional states \cite{russel_norvig,wooldridge1995intelligent}. A variety of detailed criteria also exist, such as functionalist, cognitive and legal where agency is a matter of degree rather than a binary property of systems. Different types of agentic systems satisfy agentic criteria to greater or lesser degrees. Our claim in this Position paper is that regardless of which criteria of agency is chosen, LMAs struggle to satisfy it. This is a somewhat contrarian view. LMAs appear on their face to satisfy these criteria in abundance. They seem to exhibit independence, autonomy, reactivity, reasoning capabilities and the capability to act in ways that far exceed traditional computational agents. We are able to easily and readily interact with these systems as if they were agents who can converse, generate plans and execute upon them. It is therefore understandable why the use of LLMs is often framed in terms of agentic concepts, and why it seems natural to do so. 

\subsection{Ontological Identity Conditions for Agents}
However, as we argue in this Position paper, the appearance of LMAs satisfying criteria of agency is upon deeper investigation problematic. In both philosophical and computational treatments of agency, how we identify agents and their boundaries is critical (see \cite{Olson_2024} for a review). So too is the ability to demarcate and re-identify the same agent and its properties as persisting over time. Identification of agents is generally treated as prerequisite for attributing higher-level agentic properties to a system, such as the ability to reason, deliberate, plan and execute tasks. We denote the these ontological \textit{agent identity conditions} as they consist of conditions upon the manner in which agents and their properties are identified.

\subsubsection{Identifiability}
The first condition is that LMAs be \textit{identifiable}. We must be able to (synchronically) identify what an agent is and what it is not. Identifiability requires: (i) a legible and well-formed set of criteria by which to identity an agent (\textit{identity criteria}); and (ii) a means of determining whether any system or phenomena satisfies this criteria (\textit{satisfiability criteria}). Identifiability also implies that an agent must be causally \textit{distinguishable} from its environment and other agents \cite{bennett2022b}. In other words an observer must be able to construct a means of discriminating between the environment, and those parts of that environment we call the agent: `causal identities' classifying the agent's causal interventions, and the 1st order effects of agentic properties such as autonomy, planning, reasoning, perception \cite{bennett2023c,bennettwelshciaunica2025}. 

\subsubsection{Continuity}
LMAs should maintain their (diachronic) identity over time, even a very short time \cite{Bratman2000}, throughout their instantiation. In the extreme case, an agent that exhibits no continuity from one time step to another is impossible to identify. But an agent also ought to be adaptable. It must change. An immutable system is not agentic. How continuous and which properties ought to be continuous about an agent depends on context. But it is clear that at least a de minimis degree of continuity is required to underpin anything agentic about an LMA system, including its ability to take action, their coherence and accountability. If we have continuity then an observer should, in theory, be able to construct a reasonably specific classifier or causal identity denoting a given instantiation of the agent, based upon its behaviour.  

\subsubsection{Persistence}
LMA ontology conditions also imply a degree of persistence (be it intentional \cite{dennett_intentional_1971} or psychological \cite{Parfit1984}). We distinguish persistence from continuity as the maintenance of an LMA's identity across different instantiations (distinct from a single session or instance). Persistence also implies that the properties of an agent and its identity ought not to be so sensitive to perturbations as to dissolve or radically alter. If LMAs do not exhibit persistence (e.g. where the same prompt sequencing and scaffolding lead to quite different outputs), it calls into question whether an instance of an agent was anything more than the stochastic output of the LLM itself. If we have persistence then an observer should, in theory, be able to construct a reasonably specific classifier or causal identity denoting the LMA \textit{across} instantiations, based upon its behaviour\footnote{Persistence requires a so-called `weaker' \cite{bennett2023b} causal identity than continuity.}.

\subsubsection{Consistency}
Consistency refers to the coherence of an agent’s description and actions according to which it is identified. An agent's state description - and that of any of its properties - should not be contradictorily described. It should not be described by a predicate and its negation, or by inconsistent outputs. An LMA prompt or trace riddled with contradictions would undermine most attempts to identify it as an agent. Consistency also has another sense in terms of consistency of objectives and actions with intentions. 
 
The ontological identity conditions above are not unique to agents. Nor are they sufficient to constitute something as an agent. However, they are necessary preconditions of agency. For human agents, satisfaction of identity conditions is anchored in physical embodiment and cognitive unity\cite{Parfit1984,dennett_intentional_1971}. For artificial agents which lack physical embodiment, such as corporations \cite{ListPettit2011}, it occurs by way of stable institutional practices, such as the law. For classical computational agents \cite{wooldridge_introduction_2009}, such as those based upon formal systems, it occurs via their relatively fixed ontologies of limited scope. But these necessary conditions are no longer guaranteed to be satisfied in case of LMAs. This in turn means that whether LMAs satisfy the properties that causes us to classify them as normal agents - such as reasoning, planning, autonomy, reactivity - may be called into question. This can be seen by considering how LMAs are constituted using LLMs and the structural scaffolding, such as memory, tool use and infrastructure, that supports them.

\section{LLM Pathologies} 
The foundation of an LMA is an LLM that takes prompts and outputs text (or other data modalities) conditioned on those queries. This procedure is undertaken post-training, at inference stage. While LMAs can be described at different levels of abstraction — from the micro (model) to macro (output) level - behaviour that appears recognisably agentic only arises at the macro scale. We cannot reliably identify agentic behaviour at the level of activations for example. We are therefore primarily reliant upon on macro-level outputs of LLMs and scaffolding around them to infer whether these systems meet agentic criteria and identity conditions.

Because of the way transformer models are designed \cite{radford2019language}, LLMs that form the bedrock of LMAs exhibit exhibit distinct characteristics of being stateless, stochastic, semantically sensitive and linguistically intermediated. These properties are integral to their computational power and versatility. But they also are the source of instability and uncertainty when it comes to LMA identity. For this reason and in the context of their effect on LMA identity, we denote them LLM pathologies.

\subsection{Statelessness}
First, LLMs are stateless. They do not store a persistent record of prompts and outputs from one interaction to the next \cite{merrill2024illusion,vaswani2017attention}. A query and response exist in isolation unless additional context is explicitly included. Consequently, there is no classical concept of state transition within an LLM at inference. This is in sharp distinction to traditional agents whose evolution over time is represented via changes of state. LLMs are often discussed in terms of inhering world models \cite{DBLP:journals/arxiv/abs-2305-14992,DBLP:conf/icml/NottinghamAS0H023,DBLP:journals/arxiv/abs-2306-12672,brooks_video_2024}, but this framing is problematic. LLMs do allow construction of representations of worlds, but lack the usual concept of state integral to their continuity and persistence. This ephemeral nature of LLM facilitates broad reusability: the same model can handle a plethora if different queries without being restricted to a particular context. Yet this statelessness compromises persistence of identity or memory and can give rise to inconsistencies. If the user does not reinsert context, the LLM will not recall commitments  or preferences from earlier exchanges \cite{merrill2024illusion}. This is in contrast to traditional agents which are stateful in some way (such as being described by states). Even formal computational agents usually have an identifiable system state at the level of computation (e.g. the configuration of a circuit at a point in time). 

\subsection{Stochasticity}
Second, LLMs are stochastic \cite{Bender_Gebru_McMillan-Major_Shmitchell_2021,li2023transformers,cui2024bayesian}. Repeating the same query can yield different or incorrect outputs \cite{ferrando2024do}. This unpredictability makes it difficult to distinguish stable traits that might serve as evidence of a coherent agent across time. Attempts to dampen this variability, such as adjusting temperature parameters, can have mixed results. LLMs generate output by sampling from probability distributions over tokens \cite{vaswani2017attention}. This stochasticity facilitates creativity and divergent problem-solving, but it also means repeated prompts might yield contradictory outputs. While agentic theory expects a measure of autonomy, it typically presupposes consistency over identical conditions, which is difficult to guarantee when the system’s next token is partially determined by random sampling.

\subsection{Semantic Sensitivity}
Third, LLMs exhibit high semantic sensitivity. Small perturbations in input wording can accumulate into significantly divergent outputs \cite{Wang2023,Zhu2023-vz}. Saturating an LMA with different context can alter its core properties in ways that do not occur with normal agents. This phenomenon is seen in jailbreaking, adversarial attacks \cite{mcdermott2023robustifying,moradi2021evaluating,wang2023kgpa} or contradictory responses \cite{zhang2024measuring} where slight prompt modifications can produce unexpected or inconsistent behaviour despite safeguards \cite{Mei2023}. It is also discernible in what we can denote as \textit{context attrition} \cite{shi2023large,leng2024long}, the fact that as we layer more and more context into query, the weight ascribed to features (such as agentic features) can diminish. Seemingly trivial changes in phrasing can swing an LLM’s response dramatically \cite{moradi2021evaluating,mcdermott2023robustifying}. An LLM's training on extensive text corpora can render it highly reactive to minor context shifts. Although this property can be harnessed for precise prompt engineering, it also endows LLMs with a sensitivity that can undermine the stability of its agentic properties, such as goals, reasoning, planning and execution which can all be modified by subtle textual cues.

\subsection{Linguistic Intermediation}
Fourth, LLMs are linguistically intermediated. Everything is filtered through to tokens of text and embedding-space representations \cite{shanahan2024talking}: agent definitions, environment descriptions, actions, events, and prompts. This is a form of computational dualism, meaning interaction between the LLM and the LMA's environment is filtered through an `abstraction layer' and subject to its interpretation \cite{bennett2024a}. In interactive settings this can undermine constraints upon or claims regarding behaviour. For example, the general reinforcement learning agent AIXI was initially thought to be Pareto optimal \cite{hutter2010}. This claim was undermined when the agent's performance was shown to hinge upon a choice of universal Turing machine \cite{leike2015}. Likewise, natural language constitutes an additional abstraction layer that separates the software `mind' of an LMA from the environment in which it pursues goals. All information is expressed in tokens. This is certainly nothing like how interaction occurs traditionally where agents perceive, adapt and act. Nor does it resemble how biological self-organising systems enact cognition within the world \cite{thompson2007,ciaunica2023,friston2023,bennett2025a}. Rather than sensing objects or states, the LLM receives tokenized descriptions of them and responds in kind. Some information may be lost. Adaptation is separated from embodiment by an abstraction layer, which can potentially reduce efficiency \cite{bennett2024a,bennett2025a}. Unintended ambiguities or adversarial phrasing thus can shift the LLM’s perceived reality, undermining agentic boundaries and consistent situational awareness.

\subsubsection{LLM-Only LMAs} We can see how LLM pathologies problematise LMAs by considering the simplest model of an LMA: an LLM-Only agent which operates purely via repeated calls to an underlying language model, with no cumulative context, nor external memory, nor tools. In realistic LMA scenarios, they are equipped by scaffolding, but this scenario is useful to illustrate our point. In a single prompt-response situation (with no memory and no caching as is common on LLM platforms like ChatGPT), there is minimal data to establish identity criteria upon which to identify an agent at all: only a solitary trace in the form of a query-output tuple. The boundaries of any purported agent are inseparable from this record. Individuating an agent using text data alone is difficult. It is therefore unclear how to distinguish such an agent from (i) the LLM itself, (ii) its user prompts, or (iii) the environment described by the text. 

Because the LLM is stateless, the system lacks any built-in mechanism for maintaining continuity of decisions or output over time. Nor do we see meaningful guarantees of consistency through repeated interactions. An essential element of any identity criteria is that it provides a means of being able to identify the same thing by its repeated application. But repeating the same prompt may yield different outputs. This is due to the stateless and stochastic nature of LLMs in concert with linguistic intermediation. For traditional agents, consistent responses arise because of a state which is unaffected by the act of querying. This might be physical or ontological, in the sense of classes and rules which constrain the agent. 
This is not the case for LLMs where the query can instrumentally affect the ontology of the LMA in ways very different from normal agents. 

Semantically similar inputs (e.g. schemas for designating an LMA) which vary slightly may lead to large differences in output, upon which an LMA's properties are inferred. If two queries are constructed with the intention of referring to the same agent, there is no way to determine if they refer to the same underlying LMA, or two different instantiations. From an observer's perspective, there is no `causal identity' denoting interventions by a particular agent \cite{bennett2023c,pearl2018}. Multiple LMAs could be running on top of the same LLM, making it difficult to distinguish an LMA from the LLM, or different LMAs from each other. We might try to unify multiple outputs into a coherent narrative of a single agent. But the inherent stochasticity of LLMs hinders consistency across queries. These features of LLMs challenge the identifiability of LMAs. They make it uncertain whether we can confidently infer that the same agent is persisting from one interaction to the next. 

\subsubsection{LLMs with Context} Consider the next simplest model: an LLM-only LMA but where context is added. This is a common strategy in response for achieving the semblance of persistent agency and continuous interactivity. Context includes the history or summary of previous outputs \cite{zhang2019consistent,gekhman2023robustness}. Adding context does enhance the consistency of responses across multiple prompts and enable more coherent conversations. Yet the underlying LLM remains stateless. The prompts merely carry forward relevant text from earlier exchanges. It does not solve the deeper problems of stochasticity and semantic sensitivity. Outputs remain probabilistic. Small perturbations in context can produce large and unpredictable changes in output. Consequently, any apparent persistence of an LMA is affected by how a user curates, summarises, or appends prior outputs, rather than a property of the LLM itself. If the appended history is incomplete or semantically altered, previous decisions might be lost or reversed. Text-based context can be rearranged or truncated, making it difficult to track the same agent’s boundaries across repeated interactions. Likewise, the distinguishability of LMAs is rendered uncertain when many agents share overlapping contexts. These problems undermine any strong notion of agentic continuity simply by adding context.

\section{LMA Scaffolding}
\label{section:Scaffolding}
In an attempt to overcome these limitations, architectural scaffolding can be used:
\begin{enumerate}
    \item \textit{Memory} \cite{wang_augmenting_2023,DBLP:journals/arxiv/abs-2308-01542,zhang2024survey}, such as in the form of browser caching, databases or other information registers which enable the retention of information; and
    \item \textit{Tools} \cite{schick_toolformer_2023,lu_toolsandbox_2024}, which serve as extensions that enable the LMA to effectually act via interaction with other external systems, such as via executing code, or control physical devices.
    \item \textit{Planning} \cite{huang2024understanding} as a separate and distinct module (planning is in practice manifest via a combination of memory, tool use and prompting).
    \item  \textit{Infrastructure} \cite{a16z2023emerging,chan2025infrastructure}, this may include containerised instances of LMAs e.g. via Docker \cite{docker2023llmeverywhere}, or ecosystems such as cloud technology stacks, or distributed networks.
\end{enumerate}
The purpose of scaffolding is twofold: to overcome underlying LLM pathologies and to provide LMAs with agentic capabilities, such as being able to use tools to perform tasks, or long-term memory for planning and reasoning. The modularity nature of scaffolding means there are numerous possible configurations of LMA architectures. But as we show below, despite considerable improvement in robustness, versatility and utility, scaffolding does not fully address the underlying effect of LLM pathologies in LMA identity.

\subsection{Memory Mechanisms}
Firstly, we examine the effect of memory scaffolding on LMA ontology. Consider an LMA constituted by an LLM with external memory modules or additional storage \cite{zhang2024survey}. This is a common approach in attempting to overcome the statelessness of LLMs. Memory may take the form of browser caches, external databases, or specialised vector stores that maintain relevant text, summaries of previous actions, and user interactions \cite{wang_augmenting_2023,zhong2024memorybank}. Memory mechanisms aim to instantiate a degree of persistence, allowing the LMA to reference past states or decisions. This may be to enable the LMA to perform multi-turn tasks or maintain context across longer interactions, as is required for chain-of-thought reasoning \cite{wei_chain--thought_2023}. However, the LLM itself is never truly updated by these memory modules. Memory is just another form of context. The LLM simply ingests more data as part of each query. As a result, continuity and distinguishability of LMAs rely heavily on how that memory is orchestrated. A memory store might contain a detailed record of previous interactions, but any subsequent LLM-generated output can still deviate substantially from that record. With minor prompt alterations, context attrition in long-term memory \cite{dannenhauer2023memory} or noisy retrieval, LLM outputs may contradict prior statements \cite{Mündler_He_Jenko_2023} or lose saliency, compromising the consistency of the LMA over time. 

External-memory also complicates LMA identifiability. Two or more LMAs can share the same LLM, but point to different (or partially overlapping) memory stores. In this case, it is unclear whether we have multiple distinct LMAs or just different views of the same underlying system. A single LMA can dynamically switch or shuffle memory modules depending on relevance. This can undermine its identity by causing it to become reconfigured in ways that break continuity. Memory scaffolding can improve user-facing coherence of LMAs, particularly for extended, multi-turn applications. But it falls short of guaranteeing the stable boundaries and consistent identity demanded by the traditional criteria of agency.

\subsection{Tool Use and API Integration}
The second cornerstone of LMA scaffolding is tool use and integration. When LMAs gain the ability to invoke external tools, they extend their reach into broader environments. Tools allow LLM textual outputs (e.g. code) to trigger real actions \cite{toolformer,bran2023chemcrow,DBLP:journals/arxiv/abs-2308-03427} and interact with the environment. Tool use can improve LMA performance \cite{DBLP:journals/arxiv/abs-2305-11738}. It typically requires adherence to schemas or formal inputs which can also enhance predictability. In many cases, tool use can be readily traced. If an LMA calls an API with specific parameters, we can record that event in more structured logs. This can yield a partial audit trail \cite{Waiwitlikhit2024-gv,mokander_auditing_2023}, or `trace' of actions. This is often how practical agentic applications frame LMA identification \cite{Chase-LangChain-2022,wu_autogen_2023}. LLMs can also autonomously accumulate tools when paired with actuating environments, like SDEs, for access to external services or code execution \cite{schick_toolformer_2023,lu_toolsandbox_2024}. This can include creating full applications and orchestrations among multiple applications. The agent can call an API, parse the response, and incorporate the result into its output. This leaves a considerable depth of trace data which is often used to identify LMAs and often fosters the appearance of agentic autonomy.

Yet tool use (or creation by LLMs) remains linguistically intermediated and subject to stochastic generation and semantic fragility. An inadvertent or malicious prompt can steer the model to misuse tools or produce nonsensical commands. The model’s environment is mediated entirely by language, leaving it susceptible to manipulations that exploit linguistic oversights. In any case, tool use does not solve the fundamental problems facing LMA identity. The LLM remains free to generate varying tools or contradictory tool-calling directives. Tool use may be of varying accuracy or quality \cite{lu_toolsandbox_2024,furuta2023language}. The same textual instruction that guided a prior tool operation might in another context trigger some other action, eroding consistency of the LMA's actions over time. Multiple LMAs (e.g. within the same session or instantiated in a single prompt) might share tools, further blurring where one LMA ends and another begins. The policies and tools which are used by LMAs may also be ambiguous. The action space of an LMA may be difficult to discern. Complicated coding structures integrated within the LMA itself may make tools difficult to distinguish from the LMA itself. Semantic sensitivity can mean that tool use can be subject to prompt injection \cite{zhan_injecagent_2024} and adversarial attacks in unexpected ways. As such, sophisticated tool integration does not eliminate the deeper problems of a lack of LMA continuity or how to unify the LMA into a single identifiable agent with stable boundaries.

\subsection{Cognitive Architectures and Planning Modules}
More elaborate scaffolding frameworks introduce chain-of-thought prompting, hierarchical planning modules, or meta-level reflection \cite{wei_chain--thought_2023,valmeekam_can_2023}. These can yield more systematic reasoning steps, reduce shallow guesswork, and encourage the LLM to “explain” intermediate decisions. Chains of thought are usually claimed to mirror an agent’s deliberative processes \cite{wei_chain--thought_2023}, but this is known to be problematic \cite{r93}. Despite impressive gains in reliability, these methods still revolve around the LLM and remain exposed to its pathologies. A single contradictory token can unravel the entire plan. Where standard agents update an internal state following each step, an LLM may simply produce textual placeholders of state, which, if inconsistent or corrupted, can lead to inconsistent reasoning.

These characteristics of memory integration mean that proposed cognitive architectures such as COALA \cite{sumers_cognitive_2023} which promise more elaborate internal processes for reasoning and planning (such as short-term working memory, long-term semantic and episodic stores, and procedural modules) are problematised. LMAs have no mental states per se in any traditional sense of the word. Memory scaffolding doesn't change this. An LMA's mental state is just an inference made upon using the cumulative trace of its outputs, something we infer as resembling a state of mind. 

\section{Alternative Views}
Our arguments above are premised upon a close analysis of how the features of LLMs that underpin LMAs give rise to pathological effects which propagate in ways that challenge the claim that LMAs are normal agents. Yet there are a range of alternative views and challenges to our claims, ranging from questioning whether the same criticisms might be levelled at traditional agents, to whether LLM properties such as statelessness and stochasticity may be present within normal agents also. We focus on two primary alternative views below.
\subsection{LLM pathologies are not unique}
An alternative view to the one we offer is that the problems of LMA identity are not unique to LMAs and could be said of almost any agent to some degree. For example, it may be objected that LLM weights preserve information in essentially the same way as memory and that querying via prompts approximates the process of memory itself. Responses to queries are not usually wildly chaotic. They might exhibit some deviations in edge cases, but they are reliably salient. The retention of information via weights is clearly true of LLMs. But it is currently difficult or impossible to discern at the activation exactly where or what memories reside. Information may be held in superpositions \cite{henighan2023superposition,elhage2022toy} that are difficult to disentangle and impose boundaries upon. And even then such information can still be unreliable or subject to hallucination \cite{ferrando2024do}. The reliable saliency of outputs can be relatively easily undermined. The same is true of representations of agency. LLM weights do not provide the sort of stable basis of memory we would expect of typical agents. While internal and external memory can and does improve planning abilities, LMAs can still exhibit inconsistency and unpredictability between plans and task execution \cite{mallen2023when,valmeekam_can_2023,valmeekam_large_2023}. Cognitive structuring such as COALA is therefore at base simply more intricate context architecture. It does not address the confounding of LMA identity and boundaries that arise from LLM pathologies.

\subsection{LMA identity problems are inconsequential}
A second objection to our position is that the LMA identity and ontology are largely irrelevant because all that matters is functionality. The extent to which LLM pathologies confound the identity conditions underpinning agency criteria is largely an empirical question. It may be that in the future models are developed to overcome such issues e.g. embedding statefulness in some way. In certain cases, these problems may be of limited consequence. But for complex planning tasks, especially where stakes are high, the consequences may be considerable. It may be objected that variation in how an agent is identified is inconsequential. After all, humans, corporations and other traditional agents exhibit variation in their attributes, yet their agency is not called into question. While it is true that traditional agents do indeed vary across their properties and states, the way in which they satisfy identity conditions is not grounded in anything like an LLM. A human's persistence is irrespective of how they are described. A corporation's persistence depends upon persistent reliable legal practices. A classical computational agent upon formal logically-instantiated code.


\subsection{Scaffolding can mitigate but not cure the LMA identity crisis}
Although scaffolding strategies partially mask or mitigate the pathologies, they cannot eliminate them without compromising the generative breadth that make LLMs so versatile in the first place. Memory modules, APIs, and multi-step reasoning each rely upon textual input and output; any change in assumptions encoded within scaffolding can expose underlying unpredictability. Consequently, LMAs remain constrained by the same design choices that underlie their flexible creativity. In our view, this actually highlights a feature that we conjecture is generic about LLMs and LMAs in general: that there is a necessary trade-off between the ontological stability of such systems on the one hand and their power on the other. The more ontologically rigid a system, by definition the less variance. In this sense, the conundrum of LMAs is in analogous spirit with no free lunch theorems or typical bias-variance trade-off where the more deterministic a system, the more identifiable, continuous and consistent it is, but at the cost of less expressivity, generalisability or versatility.

\section{Consequences and Responses}
\subsection{Reliability and Predictability}
There is usually little emphasis upon these foundational identity challenges faced by LMAs arising from their instantiation upon LLMs. Often uncertainty over LMAs is referred to in the small print or relegated to discussion of edge cases. The paradigm of LMAs for developers and consumers remains that of normal agents. But LMAs are not normal agents. And the effects of LLM pathologies are not edge cases.  They are inherent to any LMA architecture and it is very unclear whether they can be remedied, because they stem from transformer models on which they are based. 

\subsubsection{Interference with LMA utility}
The consequences of the unstable basis of LMA identity are significant. Without a grounding in robust, persistent identity, LMAs and any systems relying or built upon them will be subject to irreducible uncertainty. For low-risk uses this may be of little concern. But for high-stakes and high-impact decision-making, the tenuous grounding of LMAs mean that it is difficult to envisage circumstances in which the promise of truly autonomous and reliable LMA systems is achievable. LLM pathologies have been shown to confound attempts at aligning LLMs \cite{anwar2024foundational}. Attempts to govern LMAs by constructing their identity in a certain way, such as via prompt engineering constitutions \cite{DBLP:journals/arxiv/abs-2212-08073,huang_collective_2024}, imposing internal rules \cite{Schuett2024-ps} or schema, or by imposing external rules, infrastructure or environments, do not address the underlying problem of LMA identity. This lack of persistent grounding of LLMs and LMAs also manifests in well-studied behaviour such as hallucinations \cite{Azamfirei_Kudchadkar_Fackler_2023} and other problems with transformer models. From an AI safety perspective, we know little to nothing of the distribution of these failure modes. Modern model evaluation techniques being far from resembling anything like a science or possessed of the rigour expected in truly high-stakes and high-impact decision-making.

\subsubsection{Undermining trustworthiness of LMAs}
The crisis of LMA identity thus has direct pragmatic consequences for their utility and trustworthiness and the brave new world of agentic AI that relies on such assurances. LMAs may fail to deliver reliable, agentic performance on complex tasks requiring strict consistency. Over-reliance on apparent agenticness of LMAS can mask the fact that, despite sophisticated scaffolding, they may lack the persistence, continuity and predictability expected of normal agents. This interferes with their utility in production chains and workflows which require consistent, predictable, output. And it is obviously a particular problem for safety-critical applications. While scaffolding can reduce the odds of contradictory or drifting outputs, the underlying randomness and prompt sensitivity never vanish. Software integrators and domain experts must account for the possibility of erratic shifts, even after thorough testing \cite{momennejad_evaluating_2023,shavit_practices_2023}.

\subsection{A Response: Agentic Evaluations}
To address these foundational ontological dilemmas and consequences of the LMA identity crisis, our position is that researchers and developers ought to consider the following: 
\begin{itemize}
    \item \textit{Mechanistic Agentic Interpretability}. Internal to LMAs, we should aim for mechanistic interpretability tools embedded within models which can identify structures or processes occurring at the micro (model) level which correlate or causally relate to agentic properties observed of LMAs at higher levels of abstraction. Doing so would have multiple benefits. Firstly it would allow us to holistically understand in more detail how the underlying LLM architecture affects LMA agency and identity. Secondly, it ought to provide a steer towards how agents may be made more robust, or at least shed light on why this may not be possible.
    \item \textit{Agent Identity Evaluations}. External to LMAs, we ought to foster a risk-based way of quantifiably measuring and monitoring the degree of a system's \textit{agentic identity}. This we argue should include rigorous scientific evaluations of how effective LMA scaffolding configurations are at preserving the ontology and properties of agents. Agent identity evaluations could in principle be deployed at all stages of the LMA life-cycle, during LLM training, testing, inference and scaffolding stages. Quantifying the variance of agenticness using risk models would fit within how modern risk analysis works, allowing decision-makers such as boards, governments and stakeholders assess how likely agentic systems are to meet sought after criteria (e.g. reasoning, autonomy and so on).
\end{itemize}
These measures come down to assurance of AI agentic identity. They are a means of ensuring the LMA we are interacting with is actually the same LMA over time; that it actually does exhibit concrete agentic properties and that it will remain that way to an acceptable degree. Our call to re-evaluate how we approach the ontology of LMAs does not seek to address the causes of the LMA identity crisis. We hold those are intrinsic to LLMs and their versatility. However, from a practical perspective what matters is the distribution of those pathologies, failure modes and edge cases.

\section{Conclusion}
In this Position paper, we have argued that LMAs are not \textit{normal} agents, and that to deploy them most effectively we should stop treating them as if they are. While LMAs clearly demonstrate potential and can convincingly simulate agentic behaviour, they rest upon an unstable foundation presaged by the inherent pathologies of LLMs identified above. The same core characteristics that enable LLMs’ creativity and adaptability also undermine continuity of goals, autonomy, and reliability. Contemporary scaffolding approaches can patch over many surface deficiencies but as we have argued they are necessarily limited in their ability to align LMAs due to the inherent underlying computational model of LLMs themselves.

Recognising this tension from the start may encourage more pragmatic deployment strategies. Rather than presuming that LMAs are stable, we have argued that developers and researchers as a first step ought to design and apply statistical, scientifically-driven, ways of measuring the degree to which systems are agents. By doing so, the extent to which those systems - LLMs with scaffolding configurations - are ontologically robust or not can be quantified. This in turn can assist efforts in how both internal and external measures may be tuned to address these pathologies, or provide more insight into why the trade-off between LMA bias and variance is inevitable. Evaluating agentic identity also ought to provide a greater handle on how we manage the risks of using LMA systems in a way to maximise their utility and trustworthiness. With the appropriate evaluation and response, although LMAs might never be normal agents, they might not need to be.

\subsection*{Impact Statement}
Our work aims to bring a more quantitative and evidence-based methodology to AI agent research and deployment, particularly in order to heighten awareness of the risks and unique character of LMAs.

\bibliography{example_paper,library,refs-technical-governance,refs-control,refs-agentinfra,refs-examples}
\bibliographystyle{icml2025}

\end{document}